\documentclass[runningheads,a4paper]{llncs}
\usepackage{enumerate}
\usepackage{amsmath}
\usepackage{amsfonts}
\usepackage{amssymb}
\usepackage{amsbsy}
\usepackage{isomath}
\SetMathAlphabet{\mathbf}{normal}{OML}{mdput}{b}{it}
\usepackage{dsfont}
\usepackage{algorithm}
\usepackage{algorithmicx}
\usepackage{algpseudocode}
\usepackage{mathrsfs}
\usepackage{paralist}
\usepackage{epsfig}
\usepackage{subfigure}
\usepackage{makeidx} 
\usepackage{color}
\usepackage{varioref}
\usepackage{tikz}
\usepackage[sort,numbers]{natbib}

\numberwithin{equation}{section} 

\newtheorem{assumption}{Assumption}

\newcommand \E {\mathop{\mbox{\ensuremath{\mathbb{E}}}}\nolimits}

\renewcommand \Pr {\mathop{\mbox{\ensuremath{\mathbb{P}}}}\nolimits}

\newcommand{\cset}[2]{\left\{\, #1 \mathrel{:} #2 \,\right\} }
\newcommand{\meas}[1]{\lambda\left(#1\right)}

\newcommand\Reals {{\mathds{R}}}

\newcommand \CA {{\mathcal{A}}}

\newcommand \CM {{\mathcal{M}}}
\newcommand \CN {{\mathcal{N}}}

\newcommand \CP {{\mathcal{P}}}

\newcommand \CR {{\mathcal{R}}}
\newcommand \CS {{\mathcal{S}}}
\newcommand \CT {{\mathcal{T}}}

\newcommand \defn {\mathrel{\triangleq}}
\newcommand \Qf {Q_\mdp}
\newcommand \Qfp {Q_\mdp^\pol}
\newcommand \Qfs {Q_\mdp^*}

\DeclareMathAlphabet{\mathpzc}{OT1}{pzc}{m}{it}

\newcommand \GammaDist {\mathop{\mathpzc{Gamma}}\nolimits}
\newcommand \Softmax{\mathop{\mathpzc{Softmax}}\nolimits}

\newcommand \Actions {\CA}
\newcommand \States {\CS}
\newcommand \Trans {\CT}
\newcommand \tran[2] {\tau(#1 \mid #2)}

\newcommand \rew {\rho}

\newcommand \Rews {\CR}
\newcommand \disc {\gamma}
\newcommand \mdp {\mu}
\newcommand \cmp {\nu}
\newcommand \CMPs {\CN}
\newcommand \MDPs {\CM}
\newcommand \pol {\pi}
\newcommand \Pols {\CP}

\newcommand \task {m}
\newcommand \agrew {\rew_\task}
\newcommand \agpol {\pol_\task}

\newcommand \opt {\varepsilon}
\newcommand \temp {c}
\newcommand \gammaA {g_1}
\newcommand \gammaB {g_2}

\newcommand \pbel {\xi} %% policy belief
\newcommand \PBels {\mathscr{P}}
\newcommand \Pbel {\Xi}

\newcommand \hbel {\eta}
\newcommand \RBels {\mathscr{R}}
\newcommand \Rbel {\Psi}
\newcommand \rbel {\psi} %% reward belief
\newcommand \JBels {\mathscr{J}}

\newcommand \jbel {\phi} %% joint reward-policy belief
\newcommand \mbel {\omega} %% model belief
\newcommand \obel {\beta} %% optimality belief

\newcommand \loss {\ell}
\newcommand \Loss {L}

\newcommand\ind[1]{\mathop{\mbox{\ensuremath{\mathbb{I}}}}\left\{#1\right\}}

\newcommand\dd{\,\mathrm{d}}

\algblockdefx[WithProb]{WithProb}{EndProb}
[1]{{\bf w.p.} #1 {\bf do}}{\bf done}
\algcblock{WithProb}{Else}{EndProb}

\def\clap#1{\hbox to 0pt{\hss#1\hss}}

\tikzstyle{place}=[circle,draw=black,draw=blue!50,fill=blue!20,inner sep=0mm, minimum size=6mm]
\tikzstyle{hidden}=[circle,draw=black,draw=blue!50,fill=blue!01,inner sep=0mm, minimum size=6mm]
\tikzstyle{observed}=[circle,draw=black,draw=blue!50,fill=blue!10,inner sep=0mm, minimum size=6mm]
\tikzstyle{transition}=[rectangle,draw=black!50,fill=black!20,thick]

\usepackage{psfrag}

\newcommand \MTPP {\textsc{MRP}}

\newcommand \MTPPMC {\textsc{MRP-MC}}
\newcommand \MTPPMH {\textsc{MRP-MH}}
\newcommand \MTPPG {\textsc{MRP-Gibbs}}

\newcommand \MTPO {\textsc{MPO}}

\newcommand \MTPOMC {\textsc{MPO-MC}}

\newcommand \PPMH {\textsc{RP-MH}}

\newcommand \MWAL {\textsc{MWAL}}
\newcommand \IMIT {\textsc{Imit}}

\newcommand \PolWalk {\textsc{PolWalk}}
\newcommand \LinProg {\textsc{LinProg}}

\newcommand \Demos {\mathbf{D}}
\newcommand \polM {\mathbf{\pi}}
\newcommand \rewM {\mathbf{\rho}}
\newcommand \demo {\mathbf{d}_\task}
\newcommand \dmo {\mathbf{d}}

\title{Bayesian multitask inverse reinforcement learning}

\author{Christos Dimitrakakis\inst{1} \and Constantin A. Rothkopf\inst{2}}
\institute{
  EPFL, Lausanne, Switzerland\\
  \email{christos.dimitrakakis@epfl.ch} 
  \and
  Frankfurt Institute for Advanced Studies, Frankfurt, Germany\\
  \email{rothkfopf@fias.uni-frankfurt.de}
}

\toctitle{Bayesian multitask inverse reinforcement learning}
\tocauthor{C. Dimitrakakis and C. Rothkopf}

\newcommand{\keywords}[1]{\par\addvspace\baselineskip
\noindent\keywordname\enspace\ignorespaces#1}

\begin{document}
\mainmatter

\maketitle
\begin{abstract}
  We generalise the problem of inverse reinforcement learning to multiple tasks, from multiple demonstrations. Each one may represent one expert trying to solve a different task, or as different experts trying to solve the same task.  Our main contribution is to formalise the problem as statistical preference elicitation, via a number of structured priors, whose form captures our biases about the relatedness of different tasks or expert policies.  In doing so, we introduce a prior on policy optimality, which is more natural to specify. We show that our framework allows us not only to learn to efficiently from multiple experts but to  also effectively differentiate between the goals of each. Possible applications include analysing the intrinsic motivations of subjects in behavioural experiments and learning from multiple teachers.
\end{abstract}

\section{Introduction}
\label{sec:introduction}

This paper deals with the problem of multitask inverse reinforcement
learning. Loosely speaking, this involves inferring the motivations
and goals of an unknown agent performing a series of tasks in a
dynamic environment. It is also equivalent to inferring the
motivations of different experts, each attempting to solve the same
task, but whose different preferences and biases affect the solution
they choose.  Solutions to this problem can also provide principled
statistical tools for the interpretation of behavioural experiments
with humans and animals.

While both inverse reinforcement learning, and multitask learning are
well known problems, to our knowledge this is the only principled
statistical formulation of this problem.  Our first major contribution
generalises our previous work~\citep{rothkopf:dimitrakakis}, a
statistical approach for single-task inverse reinforcement learning,
to a hierarchical (population) model discussed in
Section~\ref{sec:multitask-policy-parameter-prior}.  Our second major
contribution is an alternative model, which uses a much more natural
prior on the optimality of the demonstrations, in
Section~\ref{sec:optimality-prior}, for which we also provide
computational complexity bounds. An experimental analysis of the
procedures is given in Section~\ref{sec:experiments}, while the
connections to related work are discussed in
Section~\ref{sec:discussion}. Auxiliary results and proofs are given
in the appendix.

\section{The general model}
\label{sec:model}
We assume that all tasks are performed in an environment with dynamics
drawn from the same distribution (which may be singular).  We define
the environment as a controlled Markov process (CMP) $\cmp = (\States,
\Actions, \Trans)$, with state space $\States$, action space
$\Actions$, and transition kernel $\Trans = \cset{\tran{\cdot}{s,a}}{s
  \in \States, a \in \Actions}$, indexed in $\States \times \Actions$
such that $\tran{\cdot}{s,a}$ is a probability measure\footnote{We
  assume the measurability of all sets with respect to some
  appropriate $\sigma$-algebra.}  on $\States$. The dynamics of the
environment are Markovian: If at time $t$ the environment is in state
$s_t \in S$ and the agent performs action $a_t \in A$, then the next
state $s_{t+1}$ is drawn with a probability independent of previous
states and actions:
$\Pr_\cmp(s_{t+1} \in S \mid s^t, a^t) = \tran{S}{s_t,a_t}$, 
$S \subset \States$,
where we use the convention $s^t \equiv s_1, \ldots, s_t$ and $a^t
\equiv a_1, \ldots, a_t$ to represent sequences of variables, with
$\States^t, \Actions^t$ being the corresponding product spaces.  If
the dynamics of the environment are unknown, we can maintain a belief
about what the true CMP is, expressed as a probability measure $\mbel$
on the space of controlled Markov processes $\CMPs$.

During the $\task$-th demonstration, we observe an agent acting in the
environment and obtain a $T_\task$-long sequence of actions and a
sequence of states: $\demo \defn (a^{T_\task}_\task,
s^{T_\task}_\task)$, $a^{T_\task}_\task \defn a_{\task,1}, \ldots,
a_{\task,T}$, $s^{T_\task}_\task \defn s_{\task,1}, \ldots,
s_{\task,T_\task}$.  The $\task$-th task is defined via an {\em
  unknown utility function}, $U_{\task,t}$, according to which the
demonstrator selects actions, which we wish to discover. Setting
$U_{\task,t}$ equal to the total discounted return,\footnote{Other
  forms of the utility are possible. For example, consider an agent
  who collects gold coins in a maze with traps, and where the agent's
  utility is the logarithm of the number of coins it has after it has
  exited the maze.} we establish a link with inverse
reinforcement learning:
\begin{assumption}
  The agent's utility at time $t$ is defined in terms of future
  rewards: $U_{\task,t} \defn \sum_{k=t}^\infty \disc^k r_k$,
  where $\disc \in [0,1]$ is a discount factor, and the reward $r_t$
  is given by the reward function $\agrew : \States \times \Actions \to
  \Reals$ so that $r_t \defn \agrew(s_t, a_t)$.
\end{assumption}
In the following, for simplicity we drop the subscript $\task$
whenever it is clear by context.  For any reward function $\rew$, the
controlled Markov process and the resulting utility $U$ define a
Markov decision process~\citep{Puterman:MDP:1994} (MDP), denoted by
$\mdp = (\cmp, \rew, \disc)$.  The agent uses some policy $\pol$ to
select actions $a_t \sim \pol(\cdot \mid s^t, a^{t-1})$, which
together with the Markov decision process $\mdp$ defines a
distribution\footnote{When the policy is reactive, then $\pol(a_t \mid
  s^t, a^{t-1}) = \pol(a_t \mid s_t)$, and the process reduces to
  first order Markov.} on the sequences of states, such that
$\Pr_{\mdp,\pol}(s_{t+1} \in S \mid s^t, a^{t-1}) = \int_\Actions
\tran{S}{a,s_t} \dd{\pi}(a \mid s^t, a^{t-1})$, where we use a
subscript to denote that the probability is taken with respect to the
process defined jointly by $\mdp, \pol$. We shall use this notational
convention throughout this paper.  Similarly, the {\em expected
  utility} of a policy $\pol$ is denoted by $\E_{\mdp,\pol} U_t$.  We
also introduce the family of $Q$-value functions $\cset{\Qf^\pol}{\mdp
  \in \MDPs, \pol \in \Pols}$, where $\MDPs$ is a set of MDPs, with
$\Qfp : \States \times \Actions \to \Reals$ such that: $\Qf^\pol(s, a)
\defn \E_{\mdp, \pol} \left(U_t \mid s_t = s, a_t = a\right)$.
Finally, we use $\Qfs$ to denote the optimal $Q$-value function for an
MDP $\mdp$, such that: $\Qfs(s,a) = \sup_{\pol \in \Pols} \Qfp(s,a)$,
$\forall s \in \States, a \in \Actions$.  With a slight abuse of
notation, we shall use $Q_\rew$ when we only need to distinguish
between different reward functions $\rew$, as long as the remaining
components of $\mdp$ are clear from the context.

Loosely speaking, our problem is to estimate the sequence of reward
functions $\rewM \defn \rew_1, \ldots, \rew_\task, \ldots, \rew_M$,
and policies $\polM \defn \pol_1, \ldots, \pol_\task, \ldots, \pol_M$,
which were used in the demonstrations, given the data $\Demos =
\dmo_1, \ldots, \demo, \ldots, \dmo_M$ from {\em all} demonstrations
and some prior beliefs.  In order to do this, we define a multitask
reward-policy prior distribution as a Bayesian hierarchical model.

\subsection{Multitask priors on reward functions and policies}
\label{sec:multitask-reward-policy-prior}
We consider two types of priors on rewards and policies. Their main
difference is how the dependency between the reward and the policy is
modelled. Due to the multitask setting, we posit that the reward
function is drawn from some unknown distribution for each task, for
which we assert a hyperprior, which is later conditioned on the
demonstrations.
\iftrue
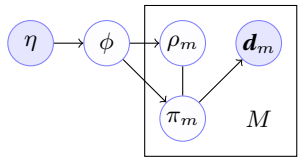
\begin{figure}[t]
  \centering
  \begin{tikzpicture}
    \node[observed] (HyperPrior) {$\hbel$};
    \node[hidden] (RewardPolicyPrior) [right of=HyperPrior] {$\jbel$};
    \node[hidden] (Reward) [right of=RewardPolicyPrior] {$\rew_\task$}; 
    \node[hidden] (Policy) [below of=Reward]{$\pol_\task$};
    % \node[hidden] (Rewards) [right of=Policy]{$\br_\task$};
    \node[observed] (Data) [right of=Reward] {$\demo$};
    \draw[->] (HyperPrior) -- (RewardPolicyPrior);
    \draw[->] (RewardPolicyPrior) -- (Reward);
    \draw[->] (RewardPolicyPrior) -- (Policy);
    \draw[-] (Reward) -- (Policy);
    %% \draw[->] (Reward) -- (Data);
    \draw[->] (Policy) -- (Data);
    \draw (1.5,0.5) rectangle (3.5, -1.5);
    \draw (3, -1) node {$M$};
  \end{tikzpicture}
  \caption{Graphical model of general multitask reward-policy
    priors. Lighter colour indicates latent variables. Here $\hbel$ is
    the hyperprior on the joint reward-policy prior $\jbel$ while
    $\rew_\task$ and $\pol_\task$ are the reward and policy of the
    $\task$-th task, for which we observe the demonstration
    $\demo$. The undirected link between $\pol$ and $\rew$
    represents the fact that the rewards and policy are jointly drawn
    from the reward-policy prior. The implicit dependencies on $\cmp$
    are omitted for clarity.}
\end{figure}
\fi
The hyperprior $\hbel$ is a probability measure on the set of joint
reward-policy priors $\JBels$. It is easy to see that, given some
specific $\jbel \in \JBels$, we can use Bayes' theorem directly to
obtain, for any $A \subset \Pols^M, B \subset \Rews^M$, where $\Pols^M, \Rews^M$ are the policy and reward product spaces:
\[
\jbel(A, B \mid \Demos) = 
\frac{\int_{A \times B} \jbel(\Demos \mid \rewM, \polM) \dd{\jbel}(\rewM, \polM)}
{\int_{\Rews^M \times \Pols^M} \jbel(\Demos \mid \rewM, \polM) \dd{\jbel}(\rewM, \polM)}
=
\prod_\task \jbel(\rew_\task, \pol_\task \mid \demo).
\]
When $\jbel$ is not specified, we must somehow estimate some
distribution on it. In the {\em empirical Bayes}
case~\citep{Robbins:EmpiricalBayes:1955} the idea is to simply find a
distribution $\hbel$ in a restricted class $H$, according to some
criterion, such as maximum likelihood.  In the {\em hierarchical
  Bayes} approach, followed herein, we select some prior $\hbel$ and
then estimate the {\em posterior distribution} $\hbel(\cdot \mid
\Demos)$.

We consider two models. In the first, discussed in
Section~\vref{sec:multitask-policy-parameter-prior}, we initially
specify a product prior on reward functions and on {\em policy
  parameters}. Jointly, these determine a unique policy, for which the
probability of the observed demonstration is well-defined. The
policy-reward dependency is exchanged in the alternative model, which
is discussed in Section~\vref{sec:optimality-prior}. There we specify a
product prior on policies and on {\em policy optimality}. This leads
to a distribution on reward functions, conditional on policies.

\section{Multitask Reward-Policy prior (\MTPP)}
\label{sec:multitask-policy-parameter-prior}
Let $\Rews$ be the space of reward functions $\rew$ and $\Pols$
the space of policies $\pol$. Let $\rbel(\cdot \mid \cmp) \in \RBels$
denote a conditional probability measure on the reward functions
$\Rews$ such that for any $B \subset \Rews$, $\rbel(B \mid \cmp)$
corresponds to our prior belief that the reward function is in $B$,
when the CMP is known to be $\cmp$. For any reward function $\rew \in
\Rews$, we define a conditional probability measure $\pbel(\cdot \mid
\rew, \cmp) \in \PBels$ on the space of policies $\Pols$.  Let
$\agrew, \agpol$ denote the $\task$-th demonstration's reward function
and policy respectively.  We use a product\footnote{Even if a prior
  distribution is a product, the posterior may not necessarily remain
  a product. Consequently, this choice does not imply the assumption
  that rewards are independent from policies.}  hyperprior\footnote{In
  order to simplify the exposition somewhat, while maintaining
  generality, we usually specify distributions on functions or other
  distributions directly, rather than on their parameters.}  $\hbel$
on the set of reward function distributions and policy distribution
$\RBels \times \PBels$, such that $\hbel(\Rbel, \Pbel) = \hbel(\Rbel)
\hbel(\Pbel)$ for all $\Rbel \subset \RBels$, $\Pbel \subset \PBels$.
Our model is specified as follows:
% \begin{subequations}
\begin{align}
  (\rbel, \pbel) & \sim \hbel(\cdot \mid \cmp), &
  \agrew \mid \rbel, \cmp & \sim \rbel(\cdot \mid \cmp),
  &
  \pol_\task \mid \pbel, \cmp, \agrew & \sim \pbel(\cdot \mid \agrew, \cmp),
  \label{eq:prior-basic}
\end{align}
% \end{subequations}
In this case, the joint prior on reward functions and policies can
be written as $\jbel(P, R \mid \cmp) \defn \int_R \pbel(P \mid \rew,
\cmp) \dd{\rbel}(\rew \mid \cmp)$ with $P \subset \Pols$, $R \subset
\Rews$, such that $\jbel(\cdot \mid \cmp)$ is a probability measure
on $\Pols \times \Rews$ for any CMP $\cmp$.\footnote{If the CMP
  itself is unknown, so that we only have a probabilistic belief
  $\mbel$ on $\CMPs$, we can instead consider the marginal $\jbel(P,
  R \mid \mbel) \defn \int_\CMPs \jbel(P, R \mid \cmp)
  \dd{\mbel}(\cmp)$.}  In our model, the only observable variables
are $\hbel$, which we select ourselves and the demonstrations $\Demos$.

%% Proof of posterior Lemma

\subsection{The policy prior}
\label{sec:policy-prior}
The model presented in this section involves restricting the policy
space to a parametric form. As a simple example, we consider
stationary soft-max policies with an inverse temperature parameter
$\temp$:
\begin{align}
  \label{eq:softmax-prior}
  \pol(a_t \mid s_t, \mdp, c)
  &=
  \Softmax(a_t \mid s_t, \mdp, \temp)
  \defn
  \frac{\exp(\temp \Qfs(s_t,a_t))}
  {\sum_a \exp(\temp \Qfs(s_t,a))},
\end{align}
where we assumed a finite action set for simplicity.  Then we can
define a prior on policies, given a reward function, by specifying a
prior $\obel$ on $\temp$.  Inference can be performed using standard
Monte Carlo methods. If we can estimate the reward functions well
enough, we may be able to obtain policies that surpass the performance
of the demonstrators.

\subsection{Reward priors}
\label{sec:reward-prior}
In our previous work~\cite{rothkopf:dimitrakakis}, we considered a
product-Beta distribution on states (or state-action pairs) for the
reward function prior.  Herein, however, we develop a more structured
prior, by considering reward functions as a measure on the state space
$\CS$ with $\rew(\CS) = 1$. Then for any state subsets $S_1, S_2
\subset \CS$ such that $S_1 \cap S_2 = \emptyset$, $\rew(S_1 \cup S_2)
= \rew(S_1) + \rew(S_2)$. A well-known distribution on probability
measures is a Dirichlet
process~\cite{Ferguson:Prior-Polya:1974}. Consequently, when $\CS$ is
finite, we can use a Dirichlet prior for rewards, such that each
sampled reward function is equivalent to multinomial parameters.  This
is more constrained than the Beta-product prior and has the advantage
of clearly separating the reward function from the $\temp$ parameter
in the policy model. It also brings the Bayesian approach closer to
approaches which bound the $L_1$ norm of the reward function such
as~\cite{syed-schapire:game-theory:nips}.

\subsection{Estimation}
The simplest possible algorithm consists of sampling directly from the
prior. In our model, the prior on the reward function $\rew$ and
inverse temperature $\temp$ is a product, and so we can simply take
independent samples from each, obtaining an approximate posterior on
rewards an policies, as shown in Alg.~\vref{alg:mc-posterior}.  While
such methods are known to converge asymptotically to the true
expectation under mild conditions~\cite{geweke1989bayesian}, stronger
technical assumptions are required for finite sample bounds, due to importance sampling in step~\ref{st:importance-sampling}.
\begin{algorithm}[h]
  \begin{algorithmic}[1]
    \For{$k=1,\ldots,K$}
    \State $\jbel^{(k)} = (\pbel^{(k)}, \rbel^{(k)}) \sim \hbel$, $\pbel^{(k)} = \GammaDist(\gammaA^{(k)}, \gammaB^{(k)})$.
    \For{$\task=1,\ldots,M$}
    %% \State $\tilde{\rew} \sim  \prod_i \Beta(\alpha_i, \beta_i)$.
    \State $\rew_\task^{(k)} \sim  \pbel(\rew \mid \cmp)$,  $\temp_\task^{(k)} \sim \GammaDist(\gammaA^{(k)}, \gammaB^{(k)})$
    \State $\mdp_\task^{(k)} = (\cmp, \disc, \rew_\task^{(k)})$, $\pol_\task^{(k)} = \Softmax(\cdot \mid \cdot, \mdp_\task^{(k)}, \temp_\task^{(k)})$, $p^{(k)}_m =  \pol_\task^{(k)} (a^T_\task \mid s^T_\task)$
    \EndFor
    \EndFor
    \State $q^{(k)} =  \prod_\task p_\task^{(k)} / \sum_{j=1}^K \prod_\task p_\task^{(j)}$ \label{st:importance-sampling}
    \State $\hat{\hbel}(B \mid \Demos) = \sum_{k=1}^K \ind{\jbel^{(k)} \in B} q^{(k)}$, for $B \subset \Rews \times \Pols$.
    \State $\hat{\rew}_\task = \sum_{k=1}^K \rew_\task^{(k)} q^{(k)}$, $\task = 1, \ldots, M$.
  \end{algorithmic}
  \caption{\MTPPMC{}: Multitask Reward-Policy Monte Carlo. Given the
    data $\Demos$, we obtain $\hat{\hbel}$, the approximate posterior on
    the reward-policy distirbution, and $\hat{\rew}_m$, the
    $\hat{\hbel}$-expected reward function for the $m$-th task.}
  \label{alg:mc-posterior}
\end{algorithm}

An alternative, which may be more efficient in practice if a good
proposal distribution can be found, is to employ a Metropolis-Hastings
sampler instead, which we shall refer to as \MTPPMH{}.  Other samplers,
including a hybrid Gibbs sampler, hereafter refered to as \MTPPG{}, such
as the one introduced in~\cite{rothkopf:dimitrakakis} are possible.

\section{Multitask Policy Optimality prior (\MTPO)}

\label{sec:optimality-prior}
Specifying a parametric form for the policy, such as the softmax, is
rather awkward and hard to justify. It is more natural to specify a
prior on the {\em optimality} of the policy demonstrated. Given the
data, and a prior over a policy class (e.g. stationary policies), we
obtain a posterior distribution on policies. Then, via a simple
algorithm, we can combine this with the optimality prior and obtain a
posterior distribution on reward functions.

As before, let $\Demos$ be the observed data and let $\pbel$ be a
prior probability measure on the set of policies $\Pols$, encoding our
biases towards specific policy types.  In addition, let $\cset{\rbel(\cdot
  \mid \pol)}{\pol \in \Pols}$ be a set of probability measures on
$\Rews$, indexed in $\Pols$, to be made precise later. In principle,
we can now calculate the marginal posterior over reward functions
$\rew$ given the observations $\Demos$, as follows:
\begin{equation}
  \rbel(B \mid \Demos) = \int_\Pols \rbel(B \mid \pol) \dd{\pbel}(\pol \mid \Demos),
  \qquad
  B \subset \Rews.
  \label{eq:optimality-posterior}
\end{equation}
The main idea is to define a distribution over reward functions, via a
prior on the optimality of the policy followed.  The first step is to
explicitly define the measures on $\Rews$ in terms of
$\opt$-optimality, by defining a prior measure $\obel$ on
$\Reals_+$, such that $\obel([0,\opt])$ is our prior that the policy
is $\opt$-optimal.  Assuming that $\obel(\opt) =
\obel(\opt \mid \pol)$ for all $\pol$, we obtain:

\begin{equation}
  \rbel(B \mid \pol) = \int_0^\infty \rbel(B \mid \opt, \pol) \dd{\obel}(\opt),
  \label{eq:optimality-policy}
\end{equation}
where $\rbel(B \mid \opt, \pol)$ can be understood as the prior
probability that $\rew \in B$ given that the policy $\pol$ is
$\opt$-optimal.
% Since $\pbel, \rbel, \obel$ are all probability measures, the marginal
% \eqref{eq:optimality-posterior} can be re-written via Fubini's
% theorem:
The marginal \eqref{eq:optimality-posterior} can now be written as:
\begin{equation}
  \rbel(B \mid \Demos) =
  \int_\Pols
  \left( 
    \int_0^\infty 
    \rbel(B \mid \opt, \pol) 
    \dd{\obel}(\opt)   
  \right)
  \dd{\pbel}(\pol \mid \Demos)
  \label{eq:optimality-posterior-expanded}
\end{equation}
We now construct $\rbel(\cdot \mid \opt, \pol)$.  Let the set of
$\opt$-optimal reward functions with respect to $\pol$ be:
$\Rews_\opt^\pol \defn \cset{\rew \in \Rews}{\|V^*_\rew -
  V^\pol_\rew\|_\infty < \opt}$.  Let $\meas{\cdot}$ be an arbitrary
measure on $\Rews$ (e.g. the counting measure if $\Rews$ is discrete).
We  can now set:
\begin{equation}
  \rbel(B \mid \opt, \pol ) \defn \frac{\meas{B \cap \Rews_\opt^\pol}}{\meas{\Rews_\opt^\pol}},
  \qquad
  B \subset \Rews.
  \label{eq:eps-optimal-reward-conditional}
\end{equation}
Then $\meas{\cdot}$ can be interpreted as an (unnormalised) prior
measure on reward functions.  If the set of reward functions $\Rews$
is finite, then a simple algorithm can be used to estimate
preferences, described below.

We are given a set of demonstration trajectories $\Demos$ and a prior
on policies $\pbel$, from which we calculate a posterior on policies
$\pbel(\cdot \mid \Demos)$. We sample a set of $K$ policies $\Pi =
\{\pol^{(i)} : i=1,\ldots,K\}$ from this posterior.  We are also given a
set of reward functions $\Rews$ with associated measure
$\meas{\cdot}$. For each policy-reward pair $(\pol^{(i)}, \rew_j) \in \Pi
\times \Rews$, we calculate the loss of the policy for the given
reward function to obtain a loss matrix:
\begin{align}
  \Loss &\defn [\loss_{i,j}]_{K \times |\Rews|},
  &
  \loss_{i,j} &\defn \sup_s V^*_{\rew_j}(s) - V^{\pol^{(i)}}_{\rew_j}(s),
  \label{eq:loss-matrix}
\end{align}
where $V^*_{\rew_j}$ and $V^{\pol^{(i)}}_{\rew_j}$ are the value
functions, for the reward function $\rew_j$, of the optimal policy and
$\pol^{(i)}$ respectively.\footnote{Again, we abuse notation slightly and
  employ $V_{\rew_j}$ to denote the value function of the MDP $(\cmp,
  \rew_j)$, for the case when the underlying CMP $\cmp$ is known.  For
  the case when we only have a belief $\mbel$ on the set of CMPs
  $\CMPs$, $V_{\rew_j}$ refers to the expected utility with respect to
  $\mbel$, or more precisely $V^\pol_{\rew_j}(s) = \E_\mbel ( U_t \mid
  s_t = s, \rew_j, \pol) = \int_{\CMPs} V^\pol_{\cmp, \rew_j}(s)
  \dd{\mbel}(\cmp)$. }

Given samples $\pol^{(i)}$ from $\pbel(\pol \mid \Demos)$, we
can estimate the integral \eqref{eq:optimality-posterior-expanded}
accurately via $\hat{\rbel}(B \mid \Demos) \defn \frac{1}{K} \sum_{i=1}^K
\int_0^\infty \rbel(B \mid \opt, \pol^{(i)}) \dd{\obel}(\opt)$.
In addition, note that the loss matrix $L$ is finite, with a number of
distinct elements at most $K \times |\Rews|$. Consequently, $\rbel(B
\mid \opt, \pol^{(i)})$ is a piece-wise constant function with respect
to $\opt$. Let $(\opt_k)_{k=1}^{K \times |\Rews|}$ be a
monotonically increasing sequence of the elements of $L$. Then
$\rbel(B \mid \opt, \pol^{(i)}) = \rbel(B \mid \opt', \pol^{(i)})$ for
any $\opt, \opt' \in [\opt_k,
\opt_{j+1}]$, and:
\begin{equation}
  \hat{\rbel}(B \mid \Demos) \defn
  \sum_{i=1}^K
  \sum_{k=1}^{K \times |\Rews|}
  \rbel(B \mid \opt_k, \pol^{(i)}) 
  \obel([\opt_k, \opt_{k+1})).
  \label{eq:sum-integral}
\end{equation}
Note that for an exponential prior with parameter $c$, we have
$\obel([\opt_k, \opt_{k+1}]) = e^{-c \opt_k} - e^{-c \opt_{k+1}}$.  We
can now find the optimal policy with respect to the expected
utility. 
\begin{theorem}
  Let $\hat{\hbel}_k(\cdot \mid \Demos)$ be the empirical posterior measure
  calculated via the above procedure and assume $\rew$ takes values in
  $[0,1]$ for all $\rew \in \Rews$. Then, for any value function
  $V_\rew$,
  \begin{equation}
    \E_\hbel (\|V_\rew -  \hat{V_\rew}\|_\infty \mid \Demos) \leq\frac{1}{(1 - \gamma)\sqrt{K}} \left(2 +
      \frac{1}{2}\sqrt{\ln K}\right),
  \end{equation}
  where the expectation is taken w.r.t the marginal distribution on
  $\Rews$.
  \label{the:mc-error}
\end{theorem}
This theorem, whose proof is in the appendix, bounds the number of
samples required to obtain a small loss in the value function
estimation, and holds with only minor modifications for both the
single and multi-task cases for finite $\Rews$. For the multi-task
case and general $\Rews$, we can use \MTPOMC{}
(Alg.~\vref{alg:mtpo-mc}), to sample $N$ reward functions from a
prior. Unfortunately the theorem does not apply directly for infinite
$\Rews$. While one could define an $\epsilon$-net on $\Rews$, and
assume smoothness conditions, in order to obtain in optimality
guarantees for that case, this is beyond the scope of this paper.
\begin{algorithm}[h]
  \begin{algorithmic}[1]
    \State Sample $N$ reward functions $\rew_1, \ldots, \rew_N \sim \rbel$.
    \For{$k=1,\ldots,K$}
    \State $(\pbel^{(k)}, \rbel^{(k)}) \sim \hbel$, where $\rbel^{(k)}$ is multinomial over $N$ outcomes.
    \For{$\task=1,\ldots,M$}
    %% \State $\tilde{\rew} \sim  \prod_i \Beta(\alpha_i, \beta_i)$.
    \State $\pol_\task^{(k)} \sim \pbel^{(k)}(\cdot \mid \demo)$.
    \EndFor
    \EndFor
    \State Calculate $\hat{\jbel}_\task(\cdot \mid \demo)$ from \eqref{eq:sum-integral} and $\{\pol_\task^{(k)} : k=1,\ldots,K\}$.
  \end{algorithmic}
  \caption{\MTPOMC{} Multitask Policy Optimality Monte Carlo posterior estimate}
  \label{alg:mtpo-mc}
\end{algorithm}

\section{Experiments}
\label{sec:experiments}
\begin{figure}[h!]
  \centering
  \subfigure[Chain task]{
    \begin{tikzpicture}
      \node[place] at (0,0) (s1) {$r_1$};
      \node[place] at (2,0) (s2) {$r_2$};
      \node[place] at (4,0) (s3) {$r_3$};
      \draw[red,->,loop above] (s1) edge (s1);
      \draw[red,->,bend right] (s2) edge (s1);
      \draw[red,->,bend right] (s3) edge (s1);
     \draw[blue,->] (s1) edge node[above]{$1 - \delta$} (s2);
      \draw[blue,->, bend right, looseness=1] (s1) edge node[below]{$\delta$} (s3);
      \draw[blue,->] (s2) edge node[above]{$\delta$} (s3);
      \draw[blue,->, loop above] (s2) edge node[below right]{$1 - \delta$} (s2);
      \draw[blue,->, loop above] (s3) edge (s3);
    \end{tikzpicture}
    \vspace{1cm}
    \label{fig:chain-graph}
  }
  \subfigure[Empirical performance gain]{
    \psfrag{difference}[B][B][0.6][0]{gain}
    \psfrag{number of runs}[B][B][0.6][0]{number of runs}
    \includegraphics[width=0.475\textwidth]{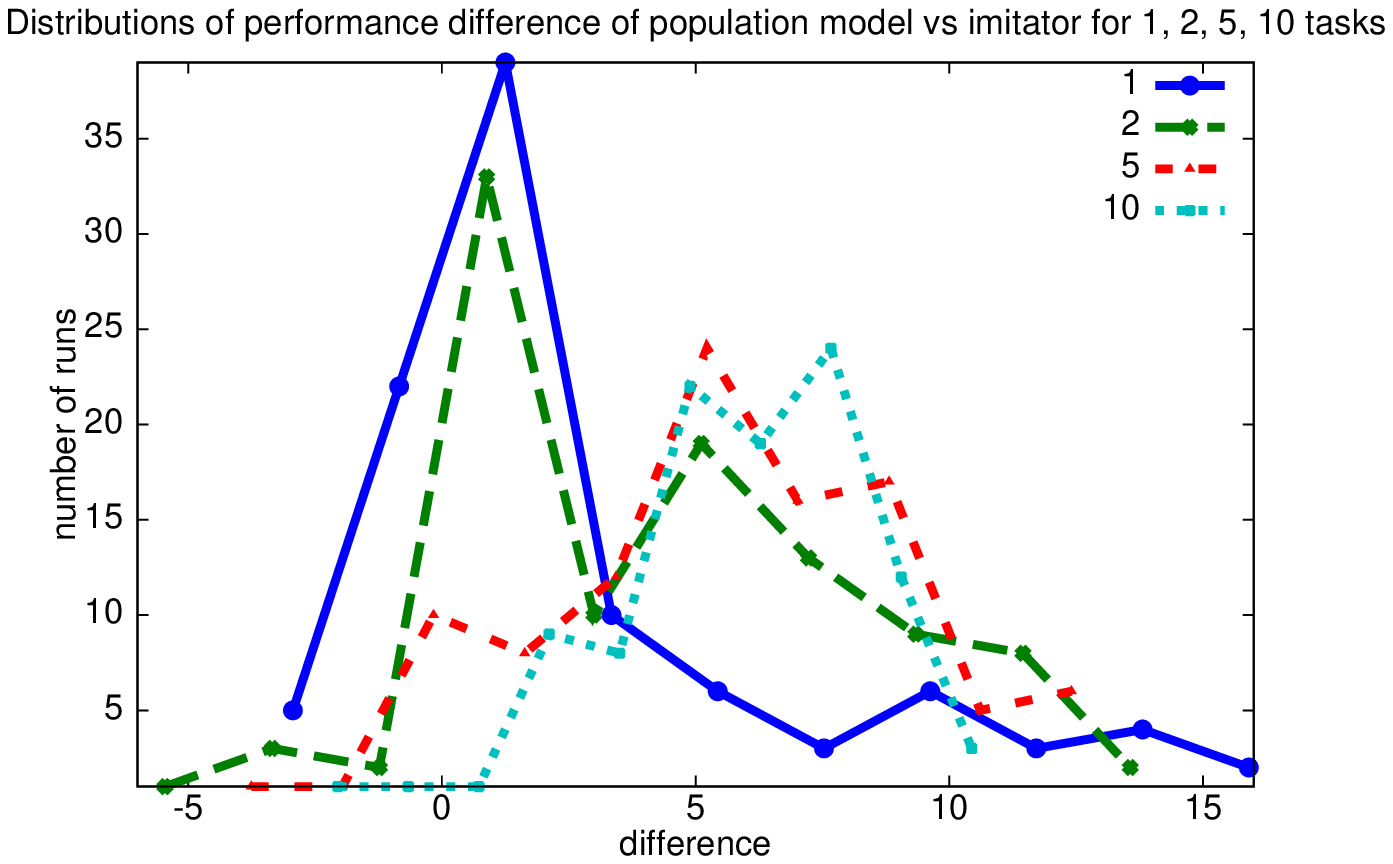}
    \label{fig:chain-diff-tasks}
  }
  \subfigure[$T=10^3$]{
    \psfrag{IMIT}[cr][][0.6][0]{\IMIT{}}
    \psfrag{MWAL}[cr][][0.6][0]{\MWAL{}}
    \psfrag{PP}[cr][][0.6][0]{\MTPPMC{}}
    \psfrag{PO}[cr][][0.6][0]{\MTPOMC{}}
    \psfrag{number of samples}[B][B][0.75][0]{number of samples/iterations}
    \psfrag{loss}[B][B][1][0]{$\loss$}
    \includegraphics[width=0.475\textwidth]{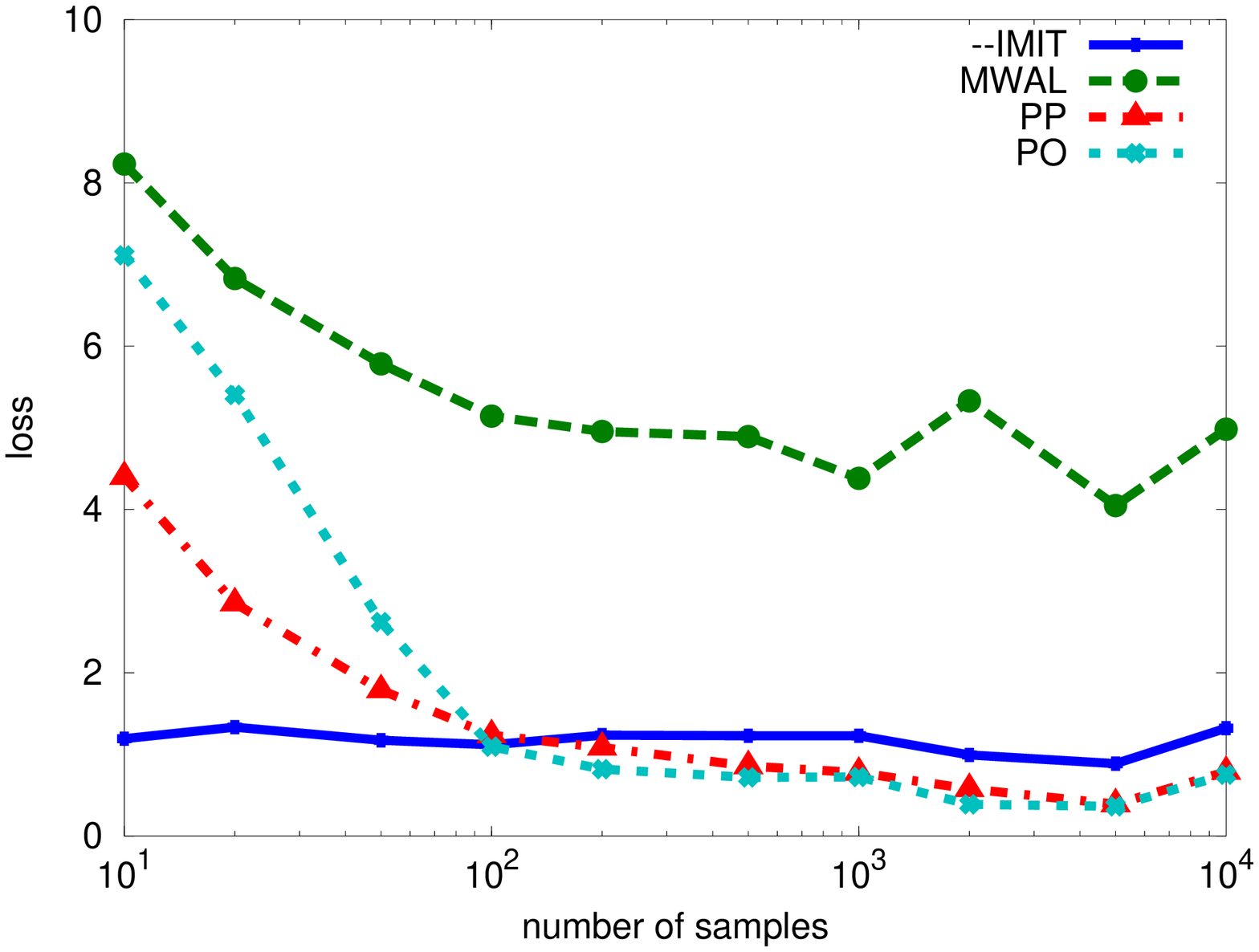}
    \label{fig:data-efficiency-a}
  }
  \subfigure[$T=10^4$]{
    \psfrag{IMIT}[cr][][0.6][0]{\IMIT{}}
    \psfrag{MWAL}[cr][][0.6][0]{\MWAL{}}
    \psfrag{PP}[cr][][0.6][0]{\MTPPMC{}}
    \psfrag{PO}[cr][][0.6][0]{\MTPOMC{}}
    \psfrag{number of samples}[B][B][0.75][0]{number of samples/iterations}
    \psfrag{loss}[B][B][1][0]{$\loss$}
    \includegraphics[width=0.475\textwidth]{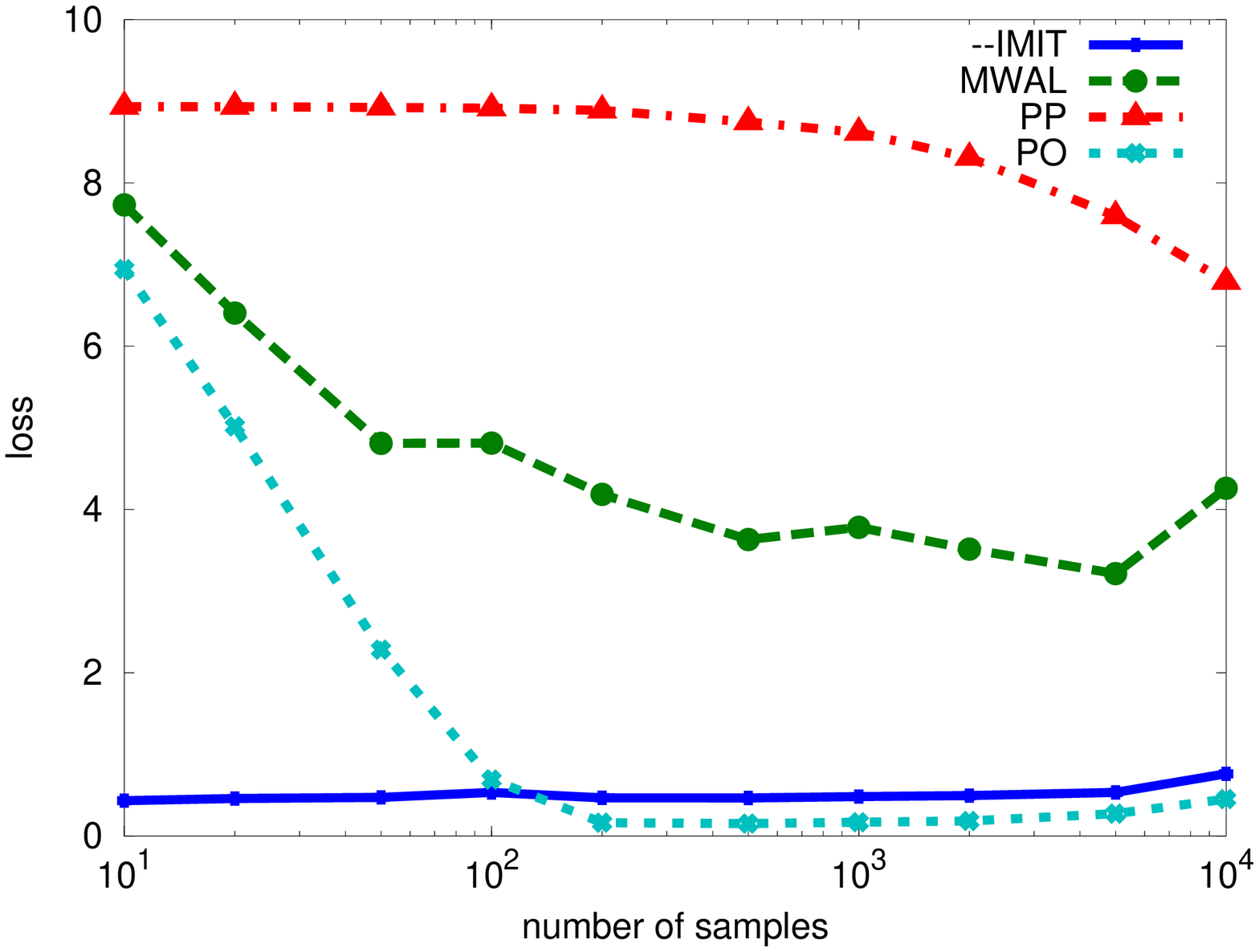}
    \label{fig:data-efficiency-b}
  }
  \caption{Experiments on the chain task. \subref{fig:chain-graph} The
    3-state version of the task. \subref{fig:chain-diff-tasks}
    Empirical performance difference of \MTPPMC{} and \IMIT{} is shown
    for $\{1, 2, 5, 10\}$ tasks respectively, with 10 total
    demonstrations. As the number of tasks increases, so does the
    performance gain of the multitask prior relative to an
    imitator. (c,d) Single-task sample efficiency in the 5-state Chain
    task with $r_1 = 0.2$, $r_2 = 0$, $r_3 = 1$. The data is
    sufficient for the imitator to perform rather well. However, while
    the $\MTPOMC{}$ is consistently better than the imitator,
     $\MTPPMC{}$ converges slowly.}
  \label{fig:data-efficiencyt}
\end{figure}

Given a distribution on the reward functions $\rbel$, and known
transition distributions, one can obtain a stationary policy that is
optimal with respect to this distribution via value iteration. This is
what single-task algorithms essentially do, but it ignores differences
among tasks. In the multi-task setting, we infer the optimal policy
$\hat{\pol}^*_\task$ for the $\task$-th task.  Its $L_1$-loss with
respect to the optimal value function is
$\loss_\task(\hat{\pol}^*_\task) \defn \sum_{s \in \CS}
V_{\rew_m}^*(s) - V_{\rew_m}^\pi(s)$. We are interested in minimising
the total loss $\sum_\task \loss_\task$ across demonstrations.
\begin{figure}[htnb]
  \centering
  \subfigure[Sampling comparison]{
    \psfrag{MH 1}[cr][][0.7][0]{\MTPPMH{} 1}
    \psfrag{MH 2}[cr][][0.7][0]{\MTPPMH{} 2}
    \psfrag{MH 4}[cr][][0.7][0]{\MTPPMH{} 4}
    \psfrag{MH 8}[cr][][0.7][0]{\MTPPMH{} 8}
    \psfrag{Monte Carlo}[cr][][0.7][0]{MTPP-MC}
    \includegraphics[width=0.47\textwidth]{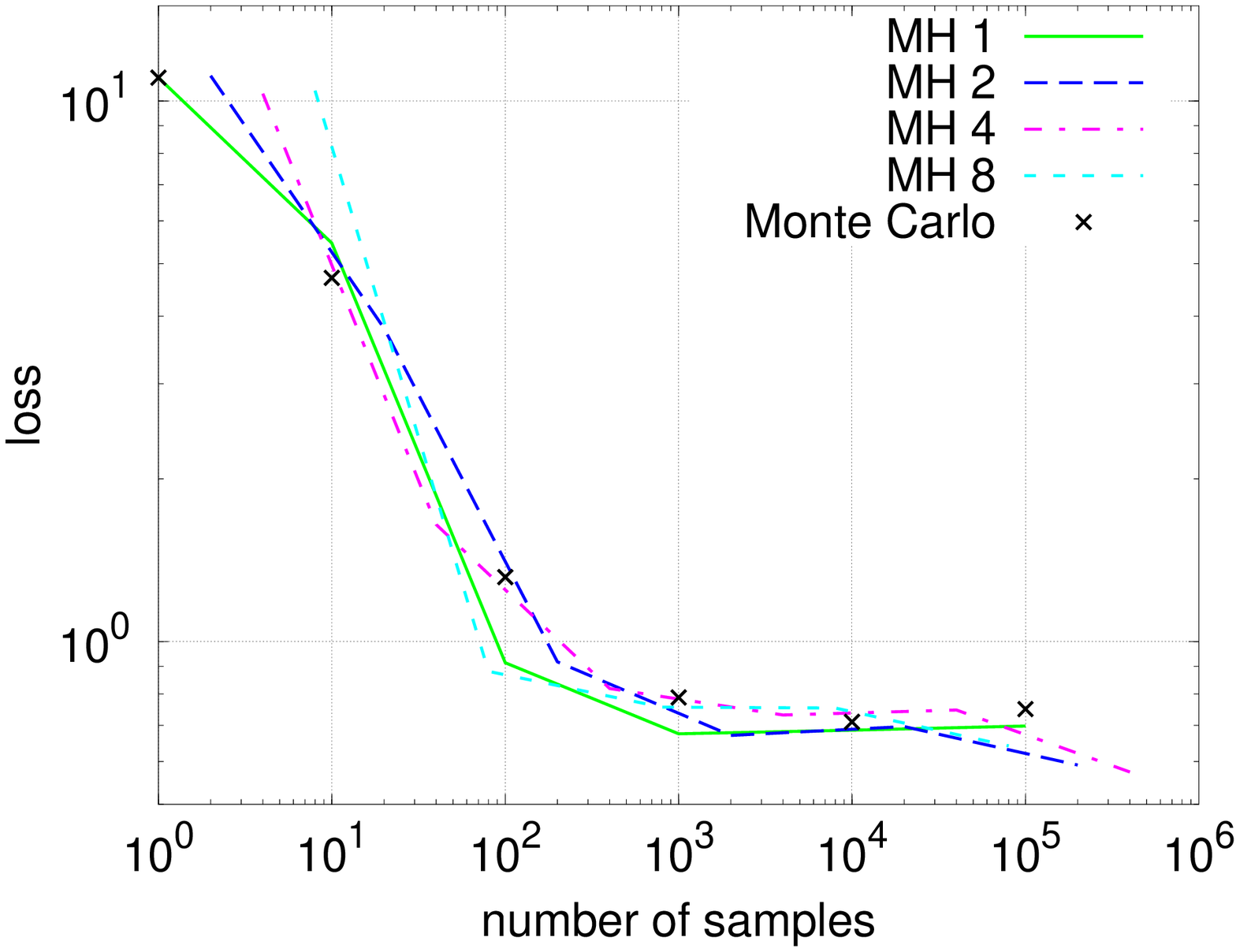}
    \label{fig:sampler-performance}
  }
  \subfigure[Model comparison]{
    \psfrag{param}[cr][][0.7][0]{\MTPPMC{}}
    \psfrag{optim}[cr][][0.7][0]{\MTPOMC{}}
    \includegraphics[width=0.47\textwidth]{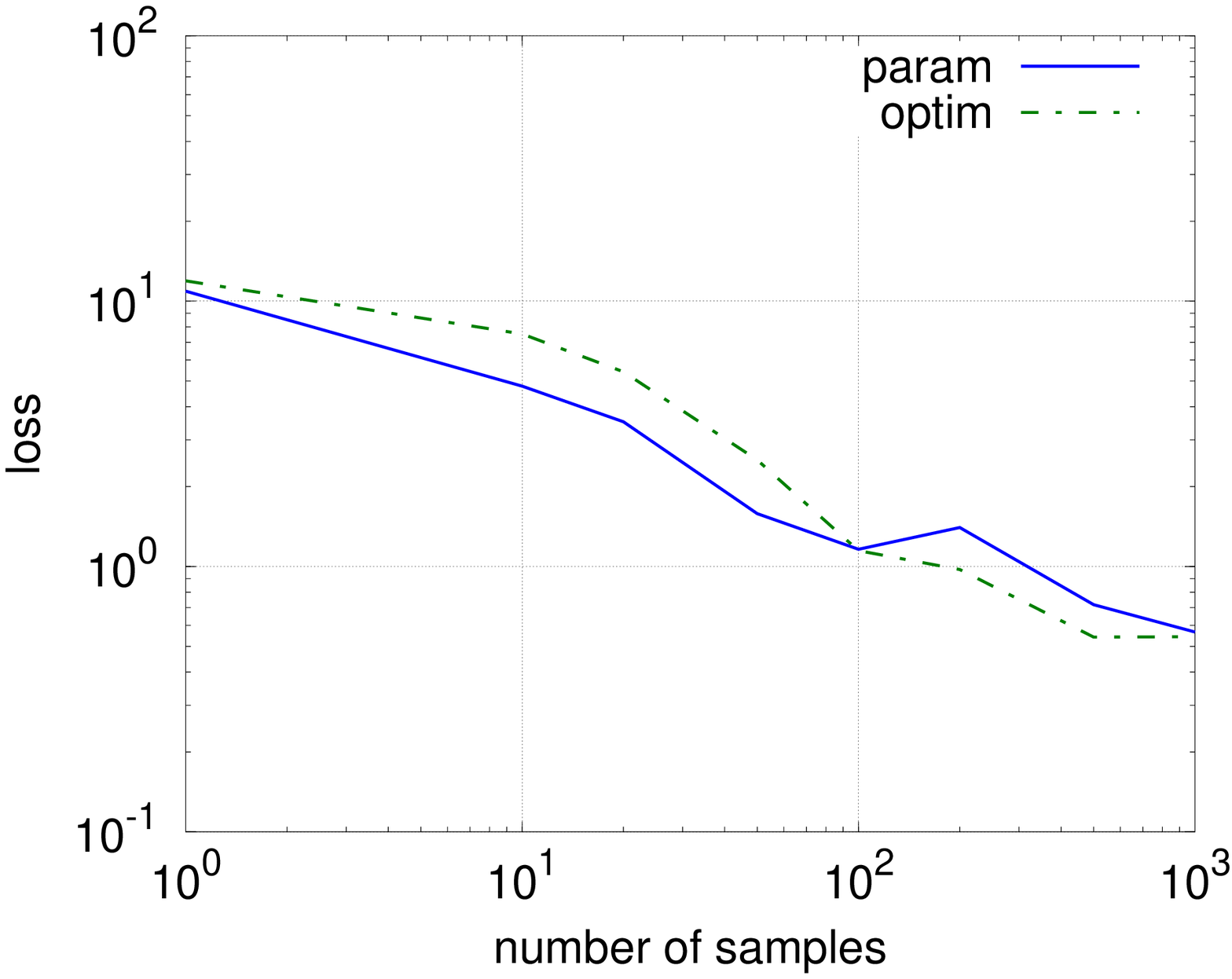}
    \label{fig:model-performance}
  }
  \caption{Expected loss for two samplers, averaged over $10^3$ runs,
    as the number of total samples increases.
    Fig.~\vref{fig:model-performance} compares the \MTPP{} and \MTPO{}
    models using a Monte Carlo estimate.
    Fig.~\vref{fig:sampler-performance} shows the performance of
    different sampling strategies for the MTPP model: {\em
      Metropolis-Hastings} sampling, with different numbers of
    parallel chains and simple {\em Monte Carlo} estimation.}
  \label{fig:preliminary-experients}
\end{figure}
We first examined the efficiency of sampling.  Initially, we used the
{\em Chain} task~\cite{dearden98bayesian} with 5 states
(c.f. Fig.~\ref{fig:chain-graph}), $\disc = 0.95$ and a demonstrator
using standard model-based reinforcement learning with
$\epsilon$-greedy exploration policy using $\epsilon = 10^{-2}$, using
the Dirichlet prior on reward functions.  As
Fig.~\vref{fig:sampler-performance} shows, for the \MTPP{} model,
results slightly favour the single chain MH
sampler. Figure~\vref{fig:model-performance} compares the performance
of the \MTPP{} and \MTPO{} models using an MC sampler.  The actual
computing time of \MTPO{} is larger by a constant factor due to the
need to calculate~\eqref{eq:sum-integral}.

In further experiments, we compared the multi-task perfomance of
\MTPP{} with that of an imitator, for the generalised chain task where
rewards are sampled from a Dirichlet prior. We fixed the number of
demonstrations to 10 and varied the nnumber of tasks. The gain of
using a multi-task model is shown in Fig.~\ref{fig:chain-diff-tasks}.
Finally, we examined the effect of the demonstration's length,
independently of the number of task.
Fig.~\ref{fig:data-efficiency-a},\ref{fig:data-efficiency-b} show that
when there is more data, then \MTPO{} is much more efficient, since we
sample directly from $\pbel(\pol \mid \Demos)$. In that case, the
\MTPPMC{} sampler is very inefficient. For reference, we include the performance of \MWAL{} and the imitator.

\begin{figure}[h!]
  \centering
  \subfigure[]{
    \psfrag{soft}[Bl][Bl][0.7][0]{soft}
    \psfrag{RP-MH}[Bl][Bl][0.7][0]{\PPMH{}}
    \psfrag{MRP-MH}[Bl][Bl][0.7][0]{\MTPPMH{}}
    \psfrag{LINPROG}[Bl][Bl][0.7][0]{\LinProg{}}
    \psfrag{POLWALK}[Bl][Bl][0.7][0]{\PolWalk{}}
    \psfrag{MWAL}[Bl][Bl][0.7][0]{\MWAL{}}
    \psfrag{Loss}[Bc][Bc][1.0][0]{$\loss$}
    \psfrag{Softmax temperature}[c][c][1.0][0]{Inverse temperature $\temp$}
    \includegraphics[width=0.475\textwidth,bb=40 15 400 300]{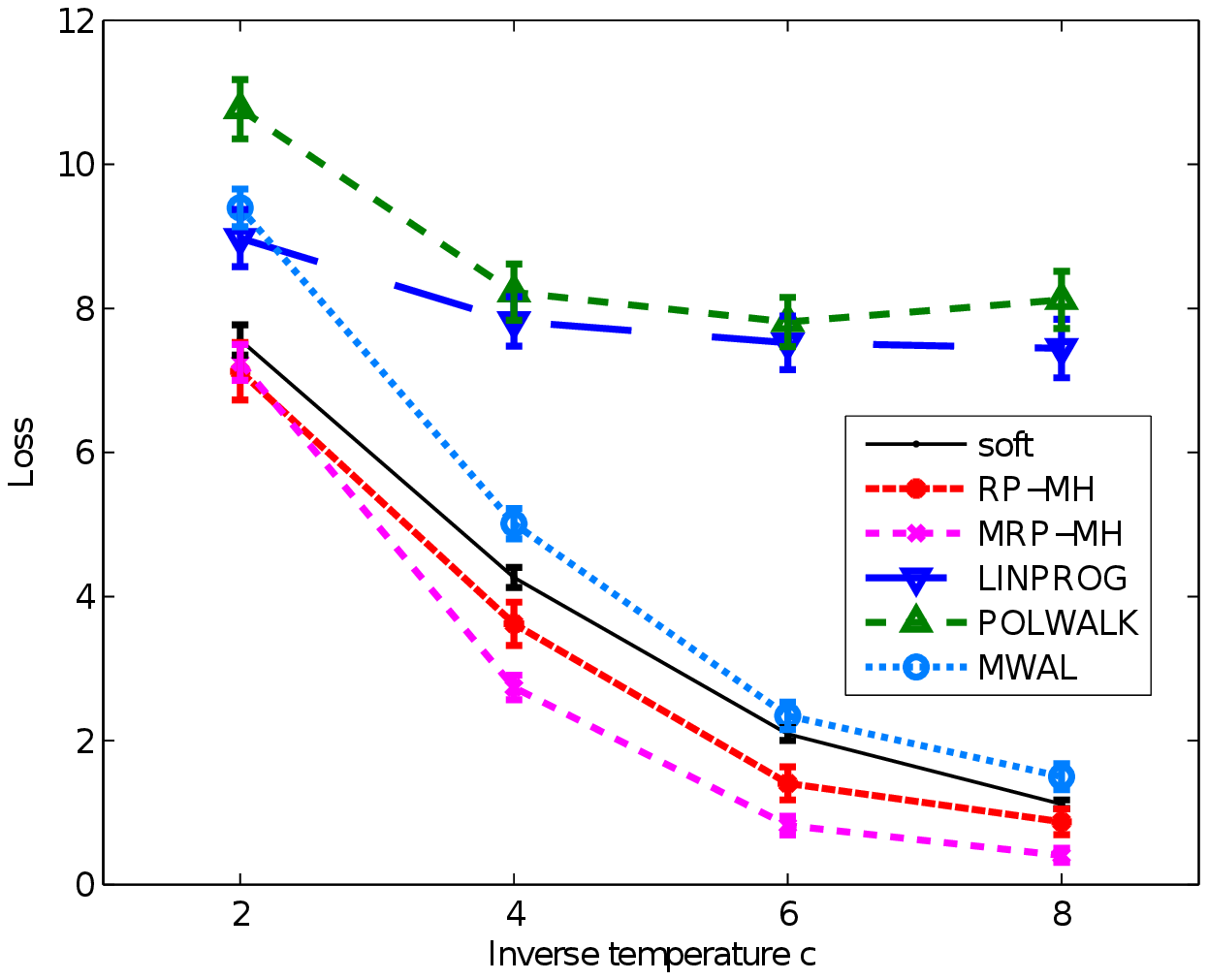}
    \label{fig:varying-temperature}
  }
  \subfigure[]{
    \psfrag{soft}[Bl][Bl][0.7][0]{soft}
    \psfrag{MH flat model}[Bl][Bl][0.7][0]{\PPMH{}}
    \psfrag{MH hierarchical}[Bl][Bl][0.7][0]{\MTPPMH{}}
    \psfrag{LP}[Bl][Bl][0.7][0]{\LinProg{}}
    \psfrag{PW}[Bl][Bl][0.7][0]{\PolWalk{}}
    \psfrag{MWAL}[Bl][Bl][0.7][0]{\MWAL{}}
    \psfrag{Loss}[Bc][Bc][1.0][0]{$\loss$}
    \psfrag{Number of tasks}[c][c][1.0][0]{Number of tasks $M$}
    \includegraphics[width=0.47\textwidth,bb=40 15 400 300]{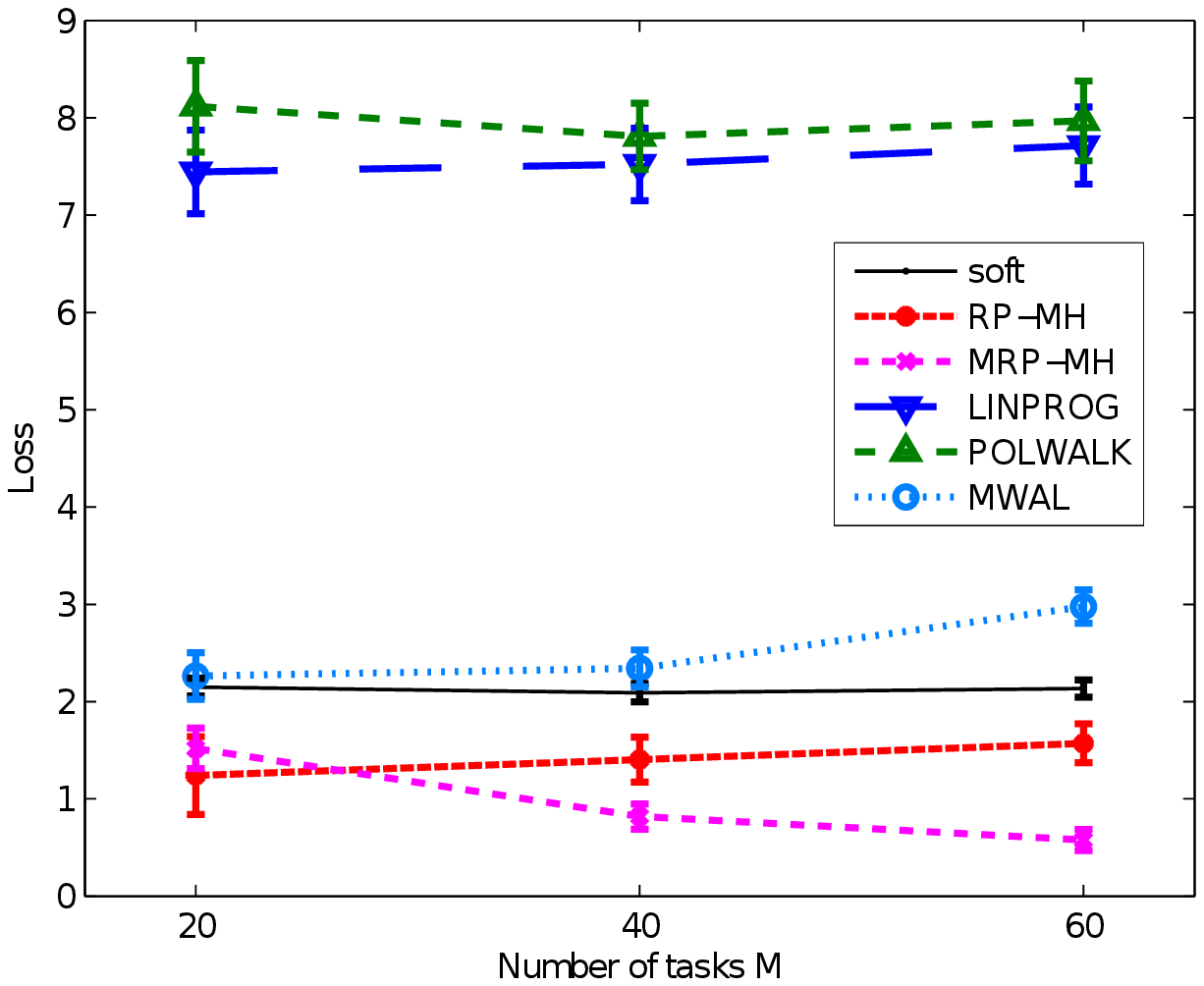}
    \label{fig:varying-tasks}
  }
  \caption{Experiments on random MDP tasks, comparing MTPP-MH with the
    original (\PPMH{}) sampler\cite{rothkopf:dimitrakakis}, a
    demonstrator employing a softmax policy (soft), Policy Walk
    (\PolWalk{})~\cite{ramachandran51bayesian} and Linear Programming
    (\LinProg{}) ~\cite{Ng00algorithmsfor}
    MWAL~\cite{syed-schapire:game-theory:nips}, averaged over $10^2$
    runs. Fig.~\vref{fig:varying-temperature} shows the loss as the
    inverse softmax temperature $\temp$ increases, for a fixed number of
    $M=20$ tasks Fig.~\vref{fig:varying-tasks} shows the loss relative
    to the optimal policy as the number of tasks increases, for fixed
    $\temp = 8$. There is one $50$-step demonstration per task. The
    error bars indicate standard error.}
  \label{fig:random-mdp}
\end{figure}

The second experiment samples {\em variants} of Random MDP
tasks~\cite{rothkopf:dimitrakakis}, from a hierarchical model, where
Dirichlet parameters are drawn from a product of $\GammaDist(1, 10)$
and task rewards are sampled from the resulting Dirichlets. Each
demonstration is drawn from a softmax policy with respect to the
current task, with $\temp \in [2, 8]$ for a total of 50 steps.  We
compared the loss of policies derived from $\MTPPMC{}$, with that of
algorithms described
in~\cite{ramachandran51bayesian,Ng00algorithmsfor,syed-schapire:game-theory:nips},
as well as a flat model~\cite{rothkopf:dimitrakakis}.
Fig.~\vref{fig:varying-temperature} shows the loss for varying
$\temp$, when the (unknown) number of tasks equals 20. While flat MH
can recover reward functions that lead to policies that outperform the
demonstrator, the multi-task model \MTPPMH{} shows a clear additional
improvement. Figure~\ref{fig:varying-tasks} shows that this increases
with the number of available demonstrations, indicating that the task
distribution is estimated well. In contrast, \PPMH{} degrades slowly,
due to its assumption that all demonstrations share a common reward
function.

\section{Related work and discussion}
\label{sec:discussion}
A number of inverse reinforcement
learning~\cite{coates2008learning,choi:inverserl-pomdp:jmlr:2011,ramachandran51bayesian,ziebart:causal-entropy,abbeel2004apprenticeship,Ng00algorithmsfor,rothkopf:dimitrakakis}
and preference
elicitation~\cite{Chu:PreferenceLearning:icml05,boutilier2002pomdp}
approaches have been proposed, while multitask learning itself is a
well-known problem, for which hierarchical Bayesian approaches are
quite natural~\citep{heskes1998solving}. In fact, two Bayesian
approaches have been considered for multitask reinforcement
learning. \citet{wilson2007multi} consider a prior on MDPs, while
\citet{lazaric2010bayesian} employ a prior on value functions.

The first work that we are aware of that performs multi-task
estimation of utilities is~\citep{birlutiu2009multi}, which used a
hierarchical Bayesian model to represent relationships between
preferences. Independently to us, \citep{Babes:ALMI:ICML:2011}
recently considered the problem of learning for multiple intentions
(or reward functions). Given the number of intentions, they employ an
expectation maximisation approach for clustering. Finally, a
generalisation of IRL to the {\em multi-agent} setting, was examined
by \citet{natarajan2010multi}. This is the problem of finding a good
{\em joint} policy, for a number of agents acting simultaneously
in the environment.

Our approach can be seen as a generalisation
of~\citep{birlutiu2009multi} to the dynamic setting of inverse
reinforcement learning; of~\citep{Babes:ALMI:ICML:2011} to full
Bayesian estimation; and of~\citep{rothkopf:dimitrakakis} to multiple
tasks. This enables significant potential applications. For example,
we have a first theoretically sound formalisation of the problem of
learning from multiple teachers who all try to solve the same problem,
but which have different preferences for doing so. In addition, the
principled Bayesian approach allows us to infer a complete
distribution over task reward functions. Technically, the work
presented in this paper is a direct generalisation of our previous
paper~\citep{rothkopf:dimitrakakis}, which proposed single task
equivalents of the policy parameter priors discussed in
Sec.~\ref{sec:multitask-policy-parameter-prior}, to the multitask
setting. In addition to the introduction of multiple tasks, we provide
an alternative policy optimality prior, which is a not only a much
more natural prior to specify, but for which we can obtain
computational complexity bounds.

In future work, we may consider non-parametric priors, such as those
considered in~\cite{NIPS2010_1188}, for the policy optimality model of
Sec.~\ref{sec:optimality-prior}. Finally, when the MDP is unknown,
calculation of the optimal policy is in general much harder. However,
in a recent paper~\cite{dimitrakakis:multi-mdp} we show how to obtain
near-optimal memoryless policies for the unknown MDP case, which would
be applicable in this setting.

\section*{Acknowledgements}
This work was partially supported by the BMBF Project ”Bernstein Fokus: 
Neurotechnologie Frankfurt, FKZ 01GQ0840”, the EU-Project IM-CLeVeR,
FP7-ICT-IP-231722, and the Marie Curie Project ESDEMUU, Grant Number
237816.

\appendix

\section{Auxillary results and proofs}
\begin{lemma}[Hoeffding inequality]%\citep{Hoeffding:SumInequalities}]
  For independent random variables $X_1, \ldots, X_n$ such that $X_i
  \in [a_i, b_i]$, with $\mu_i \defn \E X_i$ and $t>0$:
  \begin{align*}
    \Pr\left(
      \sum_{i=1}^n X_i \geq \sum_{i=1}^n \mu_i + n t
    \right) 
    &= \Pr\left(
      \sum_{i=1}^n X_i \leq \sum_{i=1}^n \mu_i - n t
    \right) 
    \leq
    \exp\left(
      -\frac{2n^2t^2}{\sum_{i=1}^n (b_i - a_i)^2}
    \right).
    %\label{eq:hoeffding}
  \end{align*}
  \label{lem:hoeffding}
\end{lemma}
\begin{corollary}
  Let $g : X \times Y \to \Reals$ be a function with total variation
  $\|g\|_{TV} \leq \sqrt{2/c}$, and let $P$ be a probability measure
  on $Y$. Define $f: X \to \Reals$ to be $f(x) \defn \int_Y g(x,y)
  \dd{P}(y)$.  Given a sample $y^n \sim P^n$, let $f^n(x) \defn
  \frac{1}{n} \sum_{i=1}^n g(x,y_i)$. Then, for any $\delta > 0$,with
  probability at least $1 - \delta$,
  $\|f - f^n\|_{\infty} < \sqrt{\frac{\ln 2/\delta}{cn}}$.
  \label{cor:function-bound}
\end{corollary}  
\begin{proof}
  Choose some $x \in X$ and define the function $h_x : Y \to [0,1]$,
  $h_x(y) = g(x, y)$. Let $h_x^n$ be the empirical mean of $h_x$ with
  $y_1, \ldots, y_n \sim P$.  Then note that the expectation of $h_x$
  with respect to $P$ is $\E h_x = \int h_x(y) dP(y) = \int g(x,y)
  dP(y) = f(x)$.  Then $ P^n\left( \left\{ y^n : \left|f(x) -
        f^n(x))\right| > t \right\} \right) < 2e^{-cnt^2}$, for any
  $x$, due to Hoeffding's inequality. Substituting gives us the
  required result.
\end{proof}
\begin{proof}[Proof of Theorem~\vref{the:mc-error}]
  Firstly, note that the value function has total variation bounded $1
  / (1 - \gamma)$. Then corollary~\vref{cor:function-bound} applies
  with $c = 2(1-\gamma)^2$. Consequently, the expected loss can be
  bounded as follows:
  \begin{align*}
    \E \|V - \hat{V}\|_\infty
    &\leq
    \frac{1}{1 -
      \gamma} \left(\sqrt{\frac{\ln 2 / \delta}{2K}} + \delta\right).
  \end{align*}
  Setting $\delta = 2/\sqrt{K}$ gives us the required result.
\end{proof}

\end{document}